\PassOptionsToPackage{sort&compress}{natbib}
\documentclass[twocolumn]{templates/arxiv/fairmeta}

\usepackage{amsmath}
\usepackage{amssymb}
\usepackage{mathtools}
\usepackage{bm}
\usepackage{algorithm}
\usepackage{algpseudocode}
\usepackage[most]{tcolorbox}
\usepackage[dvipsnames,table,xcdraw]{xcolor}
\usepackage{booktabs}
\usepackage{array}
\usepackage{tabularx}
\usepackage{ragged2e}
\usepackage{makecell}
\usepackage{multirow}
\usepackage{tikz}
\usepackage{pgfplots}
\usepackage{wrapfig}
\usetikzlibrary{pgfplots.groupplots, matrix}
\pgfplotsset{compat=1.18}
\usepackage{enumitem}
\usepackage{comment}
\usepackage{graphicx}
\usepackage{colortbl}
\usepackage{anyfontsize}
\usepackage{xspace}
\usepackage{pifont}
\usepackage{adjustbox}%

\definecolor{navyblue}{RGB}{30,130,255}
\definecolor{citecolor}{RGB}{30,130,255}
\definecolor{lightgray}{gray}{0.9}

\newcommand{\method}{RecurrentVLM\xspace}

\providecommand{\eg}{\textit{e.g.}}

\newcolumntype{L}[1]{>{\RaggedRight\arraybackslash}p{#1}}%
\newcolumntype{Y}{>{\RaggedRight\arraybackslash}X}%

\definecolor{oursbg}{RGB}{232,241,252}%
\newcommand{\oursrow}{\rowcolor{oursbg}}
\newcommand{\nr}{--}%
\newcommand{\grouprow}[2]{\multicolumn{#1}{@{}l}{\textit{#2}}\\}%

\usepackage{etoolbox}

\AtBeginDocument{\setlength{\parfillskip}{0.15\linewidth plus 0.25\linewidth}}
\newcounter{insightctr}
\tcbset{insightbase/.style={enhanced, breakable=false, boxsep=1.5pt, left=4pt, right=4pt, top=2pt, bottom=2pt,
  before skip=3.5pt, after skip=3.5pt, fontupper=\small}}

\definecolor{insc}{RGB}{38,96,168}
\newtcolorbox{insbox}{insightbase, colback=insc!5, colframe=insc, boxrule=0pt, leftrule=2.2pt, sharp corners}
\newcommand{\insight}[2]{\refstepcounter{insightctr}\begin{insbox}\textbf{\color{insc}Observation \theinsightctr.} \textbf{#1} #2\end{insbox}}
\newcommand{\obsinf}[3]{\refstepcounter{insightctr}\begin{insbox}\textbf{\color{insc}Observation \theinsightctr.} \textbf{#1}\\[1pt]\textit{Evidence.} #2\\[1pt]\textit{Inference.} #3\end{insbox}}

\newcommand{\runin}[1]{\par\vspace{5pt}\noindent\textbf{#1}}
\makeatletter
\AtBeginDocument{\pretocmd{\@hangfrom}{\parfillskip\z@ \@plus 1fil\relax}{}{\typeout{WARNING: section heading patch failed}}}
\makeatother

\setcitestyle{numbers,square,comma}
\titleformat*{\section}{\Large\sffamily\bfseries\raggedright}%
\titleformat*{\subsection}{\large\sffamily\bfseries\raggedright}
\usepackage{cuted}%
\usepackage{placeins}

\title{Recurrent Vision Language Models Solve the Shell Game}

\renewcommand{\titlelist}{{\huge\sffamily\bfseries Recurrent Vision Language Models\\Solve the Shell Game}}%

\author[1]{Dingqiang Ye}
\author[1]{Jiacong Xu}
\author[1]{Wen Zhang}
\author[1]{Yiqun Mei}
\author[1]{Vishal M. Patel}

\affiliation[1]{Johns Hopkins University}

\abstract{Tracking a hidden object through a video is effortless for humans, yet it remains a bottleneck for vision language models (VLMs). In the shell game, where a ball is hidden under one of several identical cups that are then swapped, VLMs that answer directly stay near chance, thinking models fail on long games, and fine-tuning learns only the lengths it has seen. We argue that the limit is serial computation: each swap must be composed with the last result, and a VLM of fixed depth has only a fixed number of state transitions for it. To address this, we propose \textbf{\method}, which makes every attention layer of a pretrained VLM recur over time by computing the keys and values of finished frame groups from the layer's own output. With a single answer token, \method reaches 100\% on VET-Bench and 99.6\% on MET-Bench, and 99\% on 50-swap games after training on at most 30 swaps. We also observe several interesting phenomena: length generalization emerges suddenly at an aha moment; without recurrence, each swap costs an attention layer; the ball's state forms in the first four attention layers; and recurrence keeps the latest state at the cost of details of earlier moments. These results suggest that recurrent VLMs are a potential solution for tracking the states of unobserved objects in video. Code will be released.
}

\correspondence{Dingqiang Ye}

\begin{document}

\maketitle

\begin{strip}
\centering
\includegraphics[width=\linewidth]{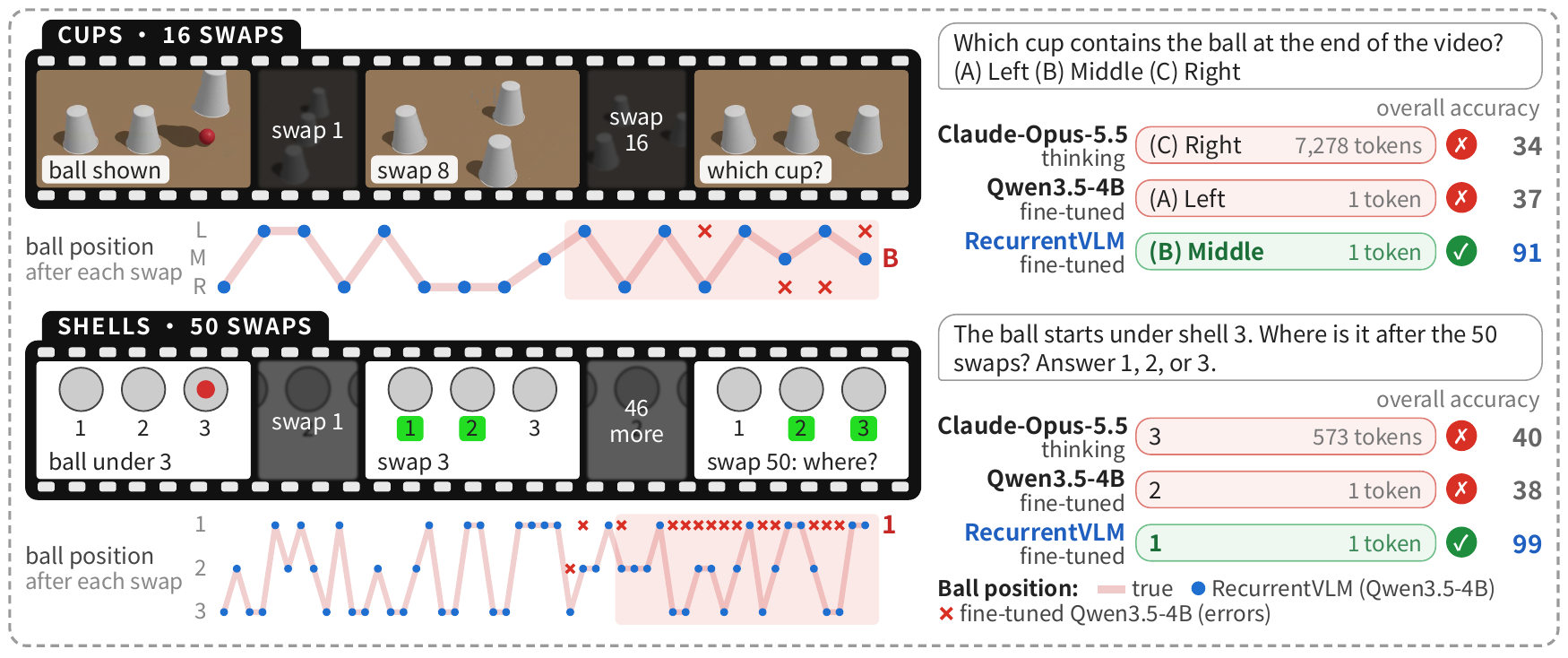}
\captionsetup{hypcap=false}
\captionof{figure}{\textbf{Current VLMs struggle with the shell game.} Claude-Opus-5.5 thinks for thousands of tokens and still errs; fine-tuned Qwen3.5-4B drifts on games longer than in training. \method stays correct at every swap with one token.}
\captionsetup{hypcap=true}
\label{fig:teaser}
\end{strip}
\setcounter{topnumber}{1}%
\setcounter{bottomnumber}{0}
\setcounter{dbltopnumber}{2}
\setcounter{totalnumber}{4}
\renewcommand{\topfraction}{0.92}
\renewcommand{\dbltopfraction}{0.92}
\renewcommand{\textfraction}{0.07}
\renewcommand{\floatpagefraction}{0.93}
\renewcommand{\dblfloatpagefraction}{0.93}
\raggedbottom
\setlength{\textfloatsep}{7pt plus 2pt minus 2pt}
\setlength{\dbltextfloatsep}{7pt plus 2pt minus 2pt}
\setlength{\floatsep}{6pt plus 2pt minus 2pt}
\setlength{\dblfloatsep}{6pt plus 2pt minus 2pt}
\setlength{\intextsep}{6pt plus 2pt minus 2pt}
\captionsetup{skip=3pt}
\captionsetup[table]{skip=3pt}
\clubpenalty=10000
\widowpenalty=10000
\section{Introduction}
\label{sec:intro}

Understanding a video often requires knowing where things are when they cannot be seen.
Humans keep track of hidden objects from infancy~\cite{baillargeon1987permanence}, and video assistants, embodied agents, and world models need the same ability to follow objects through occlusion~\cite{tokmakov2021permatrack,shin2026worldstates}.
For vision language models (VLMs), which have advanced rapidly on video benchmarks~\cite{bai2025qwen25vl,qwen3vl2025}, tracking unobserved objects is thus a basic test of how well these models understand what happens in a video.

Yet VLMs still fail this test.
The shell game is its minimal form: a ball is hidden under one of several identical cups, the cups are swapped many times, and the viewer must say where the ball is (\cref{fig:teaser}).
Since the cups look the same, no single frame shows the answer.
Answering directly, closed and open VLMs alike stay near the 33\% chance level of VET-Bench~\cite{liu2026shellgame} and MET-Bench~\cite{cohen2026metbench}.
Although they see every frame, they still cannot tell where the ball ends up after the swaps.

Existing remedies follow three routes, each with its own weakness.
Thinking models write the ball's position as text~\cite{wei2022cot,merrill2024cot}: Claude-Opus-5.5 solves the 5-swap cup games with about 1{,}000 tokens per video but drops to 34\% with 16 swaps.
Agents call external trackers~\cite{trackthenbind2026}, but the VLM itself still lacks the ability, and trackers cannot follow objects that jump between frames.
Fine-tuning Qwen3.5-4B on the task solves the short benchmark games but stays at 37\% and 38\% on 16-swap cup games and 50-swap shell games.
In short, none lets the VLM follow the ball reliably and efficiently.

This raises a question: do VLMs fail because they lack data, or because their structure lacks something?
Fine-tuning shows that data alone is not enough; the structure may be the limit, because the computation is serial: each swap permutes the cups, and the answer is the composition of all permutations.
A transformer of fixed depth performs only a bounded number of state transitions per token, too few to compose arbitrarily many permutations~\cite{merrill2024illusion,liu2023shortcuts}, and linear recurrences with eigenvalues in $[0,1]$, such as the Gated DeltaNet (GDN) layers~\cite{yang2025gdn} of Qwen3.5, cannot track permutations either~\cite{grazzi2025negative}.
A human, in contrast, keeps the ball's position in mind and updates it at every swap, a state transition over time.
This suggests that what VLMs lack is not data but such state transitions.

To test this hypothesis, we propose \textbf{\method}, which performs these state transitions inside its attention layers.
In every attention layer, \method computes the keys and values of finished frame groups from the layer's own output instead of its input, following the Recurrent Transformer~\cite{oncescu2026recurrent} but recurring once per frame group rather than once per token, so each layer passes its state forward at every frame group while the tokens of a group stay parallel.
Frame-wise supervision, which asks for the ball's position after every frame group, further helps the model learn these transitions.
Trained with the same data and losses as the fine-tuned Qwen3.5-4B, \method answers with a single token and reaches 100\% on VET-Bench and 99.6\% on MET-Bench, and keeps 91\% and 99\% on these longer games.

Detailed analyses further reveal several interesting phenomena: \emph{length generalization emerges suddenly at an aha moment; without recurrence, each swap costs an attention layer; the ball's state forms in the first four attention layers; and recurrence keeps the latest state but overwrites details of earlier moments}.

In summary, our contributions are as follows:
\begin{itemize}[leftmargin=*,topsep=2pt,itemsep=1pt]
  \item We propose \method, a simple yet effective VLM that performs state transitions over time inside its attention layers, enabling it to track unobserved objects.
  \item \method achieves state-of-the-art accuracy on VET-Bench and MET-Bench with a single answer token and generalizes to games longer than in training.
  \item We provide in-depth analyses of why recurrence works, how it emerges, and what it costs, offering insights for future VLMs that track the states of unobserved objects.
\end{itemize}

\section{Related Work}
\begin{figure*}[t]
\centering
\includegraphics[width=\textwidth]{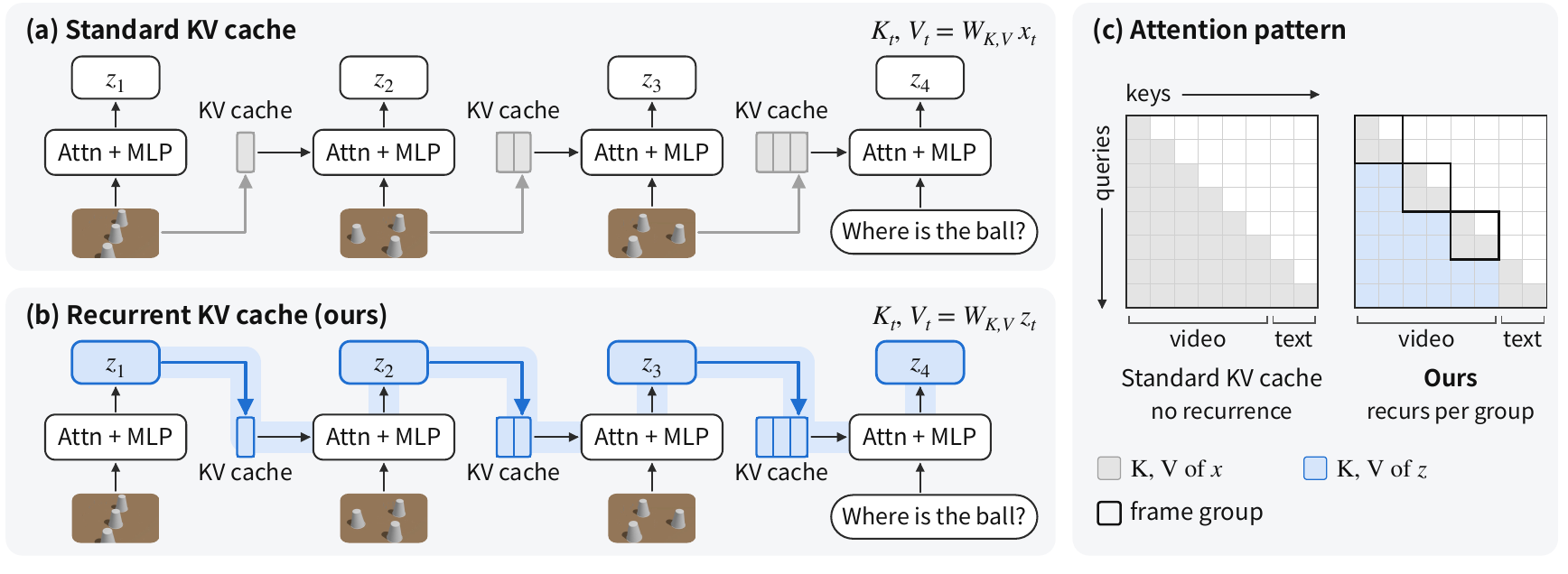}
\caption{\textbf{Recurrent KV cache.} (a) A standard attention layer of the LLM caches keys and values of its input tokens. (b) We cache keys and values of the layer's own output, once per frame group, so every attention layer becomes a recurrence over time. (c) Each query reads keys and values of earlier tokens' inputs in (a), but of earlier frame groups' outputs in (b).}
\label{fig:method}
\end{figure*}

\label{sec:related}

\runin{Video language models.}
VLMs have become the main tool for video understanding: they connect a visual encoder to a language model~\cite{alayrac2022flamingo,liu2023llava}, extend it to videos~\cite{maaz2024videochatgpt,li2025videochat,lin2024videollava,chen2024internvl,li2025llavaonevision}, and read a video as a sequence of frame tokens~\cite{bai2025qwen25vl,qwen3vl2025,gemma3_2025,hong2025glm45v,clark2026molmo2}.
Video benchmarks cover broad skills~\cite{li2024mvbench,fu2025videomme}, causal and temporal actions~\cite{xiao2021nextqa}, and questions that a single frame cannot answer~\cite{liu2024tempcompass,cores2024tvbench,shangguan2025tomato}, where TOMATO finds a 57.3-point gap between humans and the best model.
They reason with chain of thought~\cite{wei2022cot,merrill2024cot}, step supervision~\cite{nye2021scratchpad,lightman2024verify}, latent reasoning~\cite{goyal2024pause,hao2024coconut,deng2023implicit}, or agents that reason over clips and crops~\cite{liao2024videoinsta,meng2026openo3v2}.
On the shell game, SGCoT~\cite{liu2026shellgame} fine-tunes Molmo2~\cite{clark2026molmo2} to write the cup trajectory, MET-Bench trains Gemma3~\cite{gemma3_2025} with reinforcement learning, and Track, Then Bind~\cite{trackthenbind2026} calls SAM~2.
Yet all of them rely on text or external tools, whereas \method tracks hidden objects within the VLM itself.

\runin{Visual state tracking.}
Visual state tracking asks a model to follow hidden states that change over time.
Synthetic videos test compositional and causal reasoning~\cite{girdhar2020cater,yi2020clevrer}, trackers learn object permanence through occlusion~\cite{shamsian2020permanence,tokmakov2021permatrack,vanhoorick2023tcow}, the Perception Test~\cite{patraucean2023perceptiontest} films real cup games, and video world models are tested on unobserved states~\cite{shin2026worldstates}.
VET-Bench~\cite{liu2026shellgame} and MET-Bench~\cite{cohen2026metbench} hide the ball under identical cups, cards, or shells and find VLMs near chance.
Memories for long videos~\cite{song2024moviechat,chen2024videollmonline,balazevic2024mcvit} and trackers~\cite{cheng2022xmem,ravi2025sam2} store what was seen, not a state that depends on every swap.
Probes and causal interventions~\cite{alain2017probes,geiger2021causal,meng2022rome} locate board states~\cite{li2023othello,toshniwal2022chess}, entity states~\cite{li2021implicit,kim2023entity,prakash2024entity}, and permutation algorithms~\cite{li2025trackstate}, and loss drops after plateaus~\cite{power2022grokking,hoffmann2024eureka} have been traced to forming circuits~\cite{nanda2023grokking,olsson2022induction} or hidden progress~\cite{barak2022hidden}.
In contrast, this paper studies how a VLM that answers directly can learn to track such hidden states over the course of a video.

\runin{Recurrence in transformers.}
Recurrence lets a model carry a state over time.
Without it, fixed-depth transformers compute only shallow circuits per token~\cite{hahn2020limitations,merrill2023parallelism,merrill2024illusion} and learn shortcuts that break on longer sequences~\cite{liu2023shortcuts,deletang2023chomsky} unless the data format and position encoding help~\cite{anil2022length,zhou2024rasp,mcleish2024abacus}; linear recurrences share this limit~\cite{sarrof2024ssm} unless their eigenvalues may be negative~\cite{grazzi2025negative}.
Segment-level recurrence and memory~\cite{dai2019transformerxl,ding2021erniedoc,rae2020compressive,wu2022memorizing,fan2020feedback,hutchins2022blockrecurrent,bulatov2022rmt} carry hidden states across segments, fixed-size recurrences~\cite{katharopoulos2020linear,peng2023rwkv,beck2024xlstm,gu2024mamba,yang2024deltanet,yang2025gdn} replace attention, and test-time memories update their weights while reading~\cite{behrouz2025titans,zhang2025lact}.
Looped and recurrent-depth models~\cite{dehghani2019universal,fan2024looped,giannou2023looped,saunshi2025latent,graves2016act,geiping2025recurrent} repeat layers in depth instead of in time, and recurrent networks extrapolate by running more iterations~\cite{bansal2022overthinking}.
The Recurrent Transformer~\cite{oncescu2026recurrent} computes past keys and values from each layer's output.
\method extends this recurrence to video: rather than recurring once per token on text, it recurs once per frame group inside a pretrained VLM.

\section{Method}
\label{sec:method}
\begin{figure}[t]
\centering
\includegraphics[width=\linewidth]{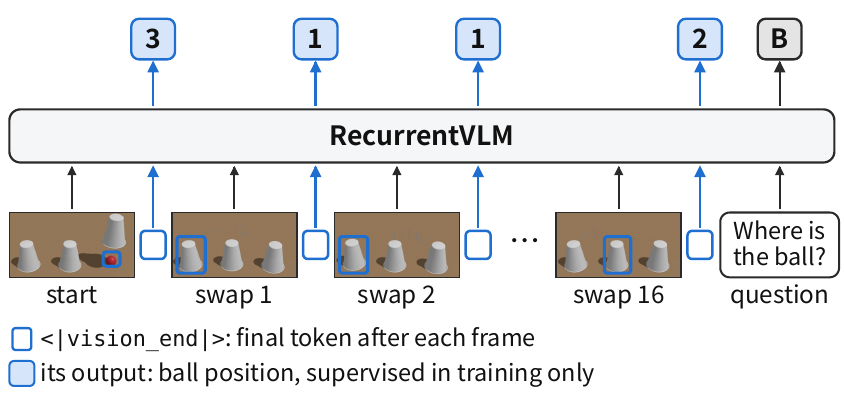}
\caption{\textbf{Frame-wise supervision.} Each frame group ends with its final token, \texttt{<|vision\_end|>}. In training, its output must name the ball's current position (boxed), and the output after the question is the answer; at test time, only the answer is generated.}
\label{fig:fws}
\vspace{2pt}%
\end{figure}

\method adds a recurrent KV cache to the attention layers of a pretrained VLM (\cref{fig:method}) and is trained with frame-wise supervision (\cref{fig:fws}).

\runin{Setting.}
A video is a sequence of frame groups $G_1, \dots, G_T$ followed by a question $q$.
A frame group holds the visual tokens of two consecutive frames, which the vision encoder of Qwen3.5 merges into one temporal patch (0.25\,s at 8\,fps); on MET-Bench, each swap image forms one group.
We build on Qwen3.5-4B, whose language model stacks 24 Gated DeltaNet (GDN) layers~\cite{yang2025gdn} and 8 attention layers, one after every three GDN layers; we number the attention layers 1 to 8.

\begin{table*}[t]
\centering
\small\setlength{\tabcolsep}{3.25pt}\renewcommand{\arraystretch}{1.0}
\caption{\textbf{Recurrence solves the shell game.} Accuracy (\%) of existing methods on VET-Bench~\cite{liu2026shellgame}, MET-Bench~\cite{cohen2026metbench}, and longer games.
Only \method stays above 90 on both longer sets. VET-Bench of Open-o3-Video, VideoINSTA and Track, Then Bind is from \citet{trackthenbind2026}, MET-Bench of the GRPO-trained Gemma3 from \citet{cohen2026metbench}; all else is ours. Baseline: Qwen3.5-4B fine-tuned on our data without recurrence. Tokens: generated per VET-Bench video. Bold: best per column; \nr: not available or not applicable.}
\label{tab:main}
\begin{adjustbox}{max width=\textwidth}
\begin{tabular}{@{}lll cc c cc c@{}}
\toprule
\multirow{2}{*}{Method} & \multirow{2}{*}{Source} & \multirow{2}{*}{Type} & \multicolumn{2}{c}{VET-Bench} & MET-Bench & \multicolumn{2}{c}{  Longer} & Cost $\downarrow$ \\
\cmidrule(lr){4-5} \cmidrule(lr){6-6} \cmidrule(lr){7-8} \cmidrule(lr){9-9}
 & & & Cup & Card & Shell & Cup$\times$16 & Shell$\times$50 & Tokens \\
\midrule
Random & -- & -- & 33.3 & 33.3 & 33.3 & 33.3 & 33.3 & -- \\
\midrule
\grouprow{9}{Closed models}
Claude-Opus-5 & Anthropic 2026 & direct answer & 32.0 & 38.0 & 36.4 & 42.0 & 32.0 & 11 \\
Claude-Opus-5.5 & Anthropic 2026 & thinking (high) & 96.0 & \textbf{100} & 97.6 & 34.0 & 40.0 & 984 \\
GPT-6 Sol & OpenAI 2026 & direct answer & 30.0 & 26.0 & 34.0 & 32.0 & 35.0 & 9 \\
GPT-6 Sol & OpenAI 2026 & thinking (high) & 84.0 & 92.0 & 99.0 & 78.0 & 97.0 & 608 \\
\midrule
\grouprow{9}{Open models}
Qwen3-VL-8B~\cite{qwen3vl2025} & Tech.\ rep.\ 2025 & direct answer & 22.0 & 28.0 & 33.0 & 34.0 & 15.0 & 549 \\
Qwen3-VL-32B~\cite{qwen3vl2025} & Tech.\ rep.\ 2025 & direct answer & 30.0 & 30.0 & 33.6 & 38.0 & 33.0 & 846 \\
Molmo2-8B~\cite{clark2026molmo2} & Tech.\ rep.\ 2026 & direct answer & 26.0 & 40.0 & 31.4 & 31.0 & \nr & 4 \\
GLM-4.6V-Flash~\cite{hong2025glm45v} & Tech.\ rep.\ 2025 & direct answer & 30.0 & 34.0 & 32.0 & 28.0 & 34.0 & 67 \\
Gemma3-4B~\cite{gemma3_2025} & Tech.\ rep.\ 2025 & direct answer & 26.0 & 28.0 & 32.8 & 26.0 & 35.0 & 6 \\
Qwen3.5-4B & Model card 2026 & direct answer & 32.0 & 40.0 & 30.8 & 41.0 & 32.0 & 6 \\
\midrule
\grouprow{9}{Fine-tuned or tool-using}
Molmo2-SGCoT~\cite{liu2026shellgame} & EMNLP 2026 & trajectory CoT, SFT & 92.0 & 90.0 & 33.6 & 70.0 & 33.0 & 368 \\
Gemma3-4B~\cite{cohen2026metbench} & ICML 2026 & direct answer, GRPO & \nr & \nr & 90.4 & \nr & \nr & \nr \\
Open-o3-Video-v2-8B~\cite{meng2026openo3v2} & arXiv 2026 & RL crop tool & 26.0 & 30.0 & 36.8 & 32.0 & 29.0 & 325 \\
VideoINSTA~\cite{liao2024videoinsta} & EMNLP-F 2024 & clip-reasoning agent & 42.0 & 28.0 & \nr & 33.3 & 43.3 & \nr \\
Track, Then Bind~\cite{trackthenbind2026} & arXiv 2026 & SAM2 tracking agent & \textbf{100} & \textbf{100} & \nr & 83.3 & \nr & 982 \\
\midrule
Baseline (Qwen3.5-4B) & Ours & direct answer, SFT & \textbf{100} & \textbf{100} & 91.6 & 37.0 & 38.0 & \textbf{1} \\
\oursrow
\textbf{\method} (Qwen3.5-4B) & Ours & direct answer, SFT & \textbf{100} & \textbf{100} & \textbf{99.6} & \textbf{91.0} & \textbf{99.0} & \textbf{1} \\
\bottomrule
\end{tabular}
\end{adjustbox}
\end{table*}

\runin{Recurrent KV cache.}
To carry a state over time, each attention layer reads its own past outputs.
Let $x_i$ and $z_i$ be the input and output of an attention layer at token $i$, where $z_i$ follows both the attention and the MLP with their residuals.
A standard layer computes keys and values from its input, $k_i = W_K \bar{x}_i$ and $v_i = W_V \bar{x}_i$, where $\bar{x}_i$ is the normalized input.
Information from a past frame therefore reaches a later token only through the inputs of deeper layers, and the number of state transitions it can undergo is bounded by the depth (\cref{fig:method}a).
Inspired by the Recurrent Transformer~\cite{oncescu2026recurrent}, once a frame group $G_s$ ends, \method recomputes its keys and values from the layer's own output with the same projections, forming the recurrent KV cache,
\begin{equation}
  \tilde{k}_j = W_K\, \bar{z}_j, \qquad \tilde{v}_j = W_V\, \bar{z}_j, \qquad j \in G_s,
  \label{eq:recurrent}
\end{equation}
keeping the rotary position of token $j$.
A query in group $G_t$ attends to the recurrent KV cache of all earlier groups and to the standard keys and values of its own group, so the output of $G_t$ depends on the outputs of earlier groups in the same layer,
\begin{equation}
  z_{G_t} = f\big(x_{G_t},\, z_{G_1}, \dots, z_{G_{t-1}}\big).
  \label{eq:recurrence}
\end{equation}
Each attention layer thus updates its state once per frame group, and the number of state transitions grows with the video, not with the depth (\cref{fig:method}b).
Within a group, tokens are still processed in parallel.

\runin{Frame-wise supervision.}
Frame-wise supervision is an auxiliary loss that helps the model learn the recurrence, since the final answer alone gives only one training signal per video.
We therefore also supervise the last token of every frame group, \texttt{<|vision\_end|>}, whose output must name the ball's current position $s_t$, \eg, 3, 1, 1, \dots, 2 in \cref{fig:fws}, while the output after the question is the answer $a$,
\begin{equation}
  \mathcal{L} = -\log p(a \mid G_{\le T}, q) - \sum\nolimits_{t=1}^{T} \log p(s_t \mid G_{\le t}).
  \label{eq:loss}
\end{equation}
The baseline uses this loss as well, so the two models differ only in the recurrent KV cache.

\begin{figure*}[t]
\centering
\includegraphics[width=\textwidth]{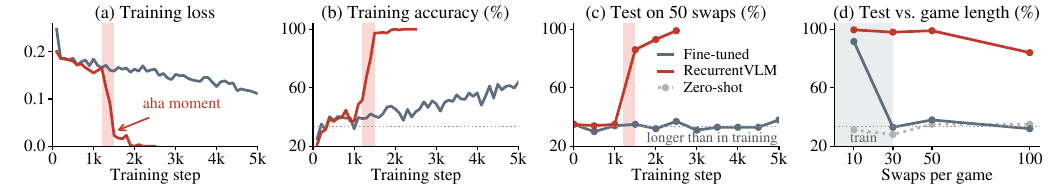}
\caption{\textbf{Length generalization emerges with an aha moment.} (a, b)~Training loss and accuracy on shell games: \method shows an abrupt drop in loss within 300 steps (shaded), an aha moment also seen in multi-step tasks~\cite{hoffmann2024eureka}, while the baseline improves slowly. (c)~Accuracy on 50 swaps, longer than in training, jumps right after the aha moment. (d)~Accuracy against the number of swaps (gray: training lengths). Fine-tuned and zero-shot: Qwen3.5-4B.}
\label{fig:length}
\end{figure*}

\runin{Training and inference.}
Training needs neither a curriculum nor full backpropagation through time.
Gradients through the recurrent KV cache stop after 8 groups, while the forward pass carries the state through all of them.
The number of swaps is sampled uniformly, so training teaches one step of the rule and the recurrence repeats it.
At inference, \method answers with a single token, so the state is never written out as text.

\section{Experiments}
\label{sec:experiments}
We answer four questions in turn: does recurrence solve the shell game (\cref{sec:exp_main}), why (\cref{sec:exp_why}), how it emerges (\cref{sec:anatomy}), and what it costs (\cref{sec:cost})?
\subsection{Setup}
\label{sec:exp_setup}
\runin{Benchmarks.}
VET-Bench~\cite{liu2026shellgame} has 50 cup and 50 card videos with 5 swaps, read at 8\,fps; MET-Bench~\cite{cohen2026metbench} has 500 shell games of 10 swap images, scored with its official prompt and parser.
For longer games, we render cup videos with 16 swaps, Cup$\times$16, with the VET-Bench generator and cut the first 100 MET-Bench test games at 30, 50, and 100 swaps, of which Shell$\times$50 is the main one.
All questions have three options (chance 33.3\%).

\runin{Training and baseline.}
Training videos come from the official VET-Bench generator (1--8 swaps of cups or cards) and the MET-Bench training split (1--30 swaps), mixed at 0.35, 0.15, and 0.5; no test video is used.
The baseline is Qwen3.5-4B fine-tuned with the same data and losses but without recurrence, trained for 5{,}000 steps as it converges more slowly; \method is trained for 2{,}500.
Both use LoRA~\cite{hu2022lora} (rank 16), learning rate $10^{-4}$ with cosine decay, and 4 videos per step on 4 RTX A5000 GPUs; all models answer directly, and unparsed answers are wrong.

\subsection{Main results}
\begin{table}[t]
\centering
\small\setlength{\tabcolsep}{3.3pt}\renewcommand{\arraystretch}{1.0}
\caption{\textbf{More cups and more balls.} Accuracy (\%) on our MET-Bench-style shell games with 3 or 5 cups and 1--3 balls, 100 games each (random: 33.3 and 20.0); both models train on the same mix of all settings, with up to 30 swaps per training game.}
\label{tab:multiball}
\begin{tabular}{@{}cl cccccc@{}}
\toprule
\multirow{2}{*}[-3pt]{Swaps} & \multirow{2}{*}[-3pt]{Model} & \multicolumn{3}{c}{3 cups, balls} & \multicolumn{3}{c}{5 cups, balls} \\
\cmidrule(lr){3-5}\cmidrule(l){6-8}
 & & 1 & 2 & 3 & 1 & 2 & 3 \\
\midrule
\multirow{2}{*}{30} & Qwen3.5-4B (fine-tuned) & 90 & 91 & 94 & 95 & 93 & 90 \\
 & \method & \textbf{100} & \textbf{100} & \textbf{100} & \textbf{96} & \textbf{98} & \textbf{93} \\
\midrule
\multirow{2}{*}{50} & Qwen3.5-4B (fine-tuned) & 38 & 44 & 32 & 29 & 25 & 20 \\
 & \method & \textbf{100} & \textbf{99} & \textbf{100} & \textbf{87} & \textbf{92} & \textbf{92} \\
\bottomrule
\end{tabular}
\end{table}

\begin{table*}[t]
\centering
\small\setlength{\tabcolsep}{4pt}\renewcommand{\arraystretch}{1.0}
\caption{\textbf{Different recurrence mechanisms.} Accuracy (\%) on the benchmarks and longer games of \cref{tab:main}. Each row adds one mechanism to the baseline, Qwen3.5-4B (fine-tuned), trained with the same data and losses. \textit{Recurrent state}: what is carried over time beyond the KV cache; \textit{Train time}: per step, relative to the baseline; per-token recurrence is about 50$\times$ slower and not trained (--).}
\label{tab:mechanisms}
\begin{adjustbox}{max width=\textwidth}
\begin{tabular}{l l l c cc cc}
\toprule
\multirow{2}{*}{Method} & \multirow{2}{*}{Applied to} & \multirow{2}{*}{Recurrent state} & \multirow{2}{*}{\makecell{Train\\time}} & \multicolumn{2}{c}{Benchmarks} & \multicolumn{2}{c}{Longer games} \\
\cmidrule(lr){5-6} \cmidrule(l){7-8}
 & & & & VET & MET & Cup$\times$16 & Shell$\times$50 \\
\midrule
Random & -- & -- & -- & 33.3 & 33.3 & 33.3 & 33.3 \\
Qwen3.5-4B (fine-tuned) & -- & GDN, eigenvalues in $[0,1]$ & 1.0$\times$ & 100 & 91.6 & 37.0 & 38.0 \\
\midrule
\quad + Looped Transformer~\cite{dehghani2019universal,fan2024looped} & whole model, twice & none & 1.8$\times$ & 100 & 57.4 & 31.0 & 42.0 \\
\quad + Negative-eigenvalue GDN~\cite{grazzi2025negative} & 24 GDN layers & GDN, eigenvalues in $[-1,1]$ & 1.0$\times$ & 94.0 & 98.8 & 30.0 & 49.0 \\
\quad + Recurrent Memory Transformer~\cite{bulatov2022rmt} & input tokens & memory tokens & 6.8$\times$ & 36.0 & 69.6 & 35.0 & 34.0 \\
\quad + Recurrent Transformer~\cite{oncescu2026recurrent} & 8 attention layers & layer outputs, per token & ${\sim}$50$\times$ & \nr & \nr & \nr & \nr \\
\oursrow
\quad + \textbf{\method} & 8 attention layers & layer outputs, per frame group & 2.5$\times$ & 100 & 99.6 & 91.0 & 99.0 \\
\bottomrule
\end{tabular}
\end{adjustbox}
\end{table*}

\label{sec:exp_main}
We test \method against existing methods, on longer games, and with more cups and balls.

\runin{Recurrence solves the shell game.}
We compare \method with closed, open, fine-tuned, and tool-using models on VET-Bench, MET-Bench, and the longer games (\cref{tab:main}).
Models that answer directly only guess, with 15--42\% on every set, around the 33\% chance level.
Thinking reaches 84--100\% on VET-Bench but pays with hundreds of tokens per video: Claude-Opus-5.5 still falls to 34\% and 40\% on the longer games, and GPT-6 Sol holds up at 78\% and 97\% with 608 tokens.
Track, Then Bind~\cite{trackthenbind2026} solves VET-Bench with tool calls but not shell games.

After supervised fine-tuning, the baseline learns the short games, with 100\% on VET-Bench and 91.6\% on MET-Bench, but stays at 37\% and 38\% on Cup$\times$16 and Shell$\times$50.
\method reaches 99.6\% on MET-Bench and 91\% and 99\% on these longer games.
After fine-tuning, \method thus solves both benchmarks and their longer games with a direct, single-token answer.

\runin{Length generalization.}
Solving the test games could still mean fitting their lengths, so we test up to three times the training length.
Trained on up to 30 shell swaps, \method reaches 98\%, 99\%, and 84\% with 30, 50, and 100 shell swaps (\cref{fig:length}d).
The baseline does not even fit its own training range: it is near chance at 30 shell swaps (\cref{fig:length}d) and on Cup$\times$16 (\cref{tab:main}).
Recurrence extrapolates to longer games, and fine-tuning does not.

\begin{figure*}[t]
\centering
\includegraphics[width=\textwidth]{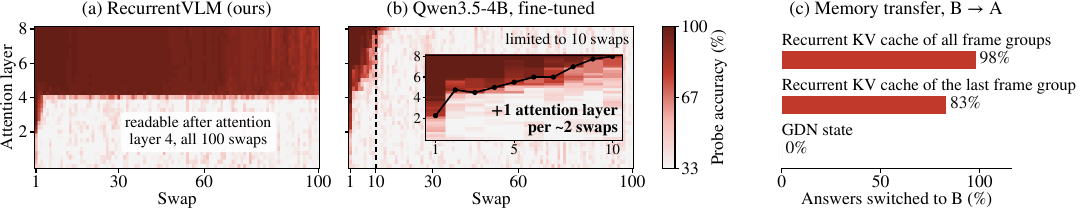}
\caption{\textbf{Probing the ball across attention layers.} (a, b)~Linear probes reading the ball position from every layer output (three GDN layers between attention layers) after each swap of 100-swap games, fit on the first 30 (inset: swaps 1--10). \method reads the ball after attention layer 4 at all swaps, while the baseline needs one more attention layer per about two swaps, which limits it to 10 swaps. (c)~Memory transfer: midway through game A, we copy part of game B's recurrent state and count answers switching to B.}
\label{fig:mechanism}
\end{figure*}

\runin{More cups and more balls.}
A single ball is the simplest state, so we also hide 1--3 balls under 3 or 5 cups, ask about one ball after the video, and supervise every ball after each swap (\cref{tab:multiball}).
Trained only on these shell games, with twice the shell data of \cref{tab:main}, both models answer at least 90\% up to the 30 swaps seen in training.
At 50 swaps, the baseline falls to 20--44\%, while \method keeps 87--100\%.
Recurrence also extrapolates with more cups and more balls.

\insight{Recurrence learns the rule.}{Direct answers guess, thinking pays with hundreds of tokens, and fine-tuning fits only the lengths it has seen. Recurrence carries what it learns to longer games and more objects.}

\subsection{Why does recurrence solve the shell game?}
\begin{figure}[t]
\centering
\includegraphics[width=\linewidth]{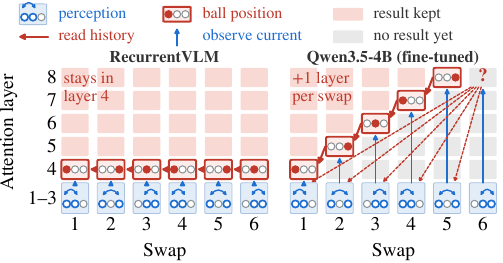}
\caption{\textbf{Our hypothesis of why depth runs out.} Each swap is combined with the last result. \method keeps it in one layer; the baseline reads earlier swaps one layer lower, so the result climbs and swap 6 finds only raw swaps (dashed).}
\label{fig:handover}
\end{figure}

\label{sec:exp_why}
We mainly explore this question with the following three experiments.
First, we compare different recurrence mechanisms on the shell game.
Second, we probe the ball position layer by layer across the attention layers.
Third, we observe how the training loss of both models changes.

\runin{Different recurrence mechanisms.}
We add each mechanism to the baseline and train it with the same data and losses (\cref{tab:mechanisms}).
Running the model twice, as a Looped Transformer~\cite{dehghani2019universal,fan2024looped}, adds depth but no state transitions over time and reaches only 57.4\% on MET-Bench.
Negative eigenvalues let the GDN layers~\cite{grazzi2025negative} solve MET-Bench (98.8\%) but not the longer games (30\% and 49\%).
Memory tokens passed between frame groups~\cite{bulatov2022rmt} cost 6.8$\times$ per step and stay at chance on VET-Bench and the longer games; per-token recurrence~\cite{oncescu2026recurrent} would be 50$\times$ slower.

Recurrence over frame groups reaches the longer games at 2.5$\times$ the cost per step and with half the steps.
More depth is thus no substitute for state transitions over time.

\runin{Probing the ball across attention layers.}
To see where each model holds the ball, we train a linear probe, a linear classifier on the output of one layer, to read the ball position after each swap of 100-swap shell games, for every layer of both models; the probes are fit on the first 30 swaps and tested on all 100 (\cref{fig:mechanism}a,\,b).

Interestingly, we observe that \emph{in the baseline, the first layer from which the ball can be read moves one attention layer deeper about every two swaps} (correlation 0.96), and after about ten swaps no layer is left.
In \method, the ball is readable after attention layer 4 at all 100 swaps (94--96\% on average, 87\% over the last 10), and its per-swap accuracy, the accuracy of its frame-wise answers, is 95.6\%, against only 41.7\% for the baseline.

We hypothesize that, without recurrence, each swap is combined with the last result one attention layer higher (\cref{fig:handover}).
Each swap thus costs the baseline an attention layer, while recurrence keeps the ball in the same attention layer at every one of the 100 swaps.

\begin{figure*}[t]
\centering
\includegraphics[width=\textwidth]{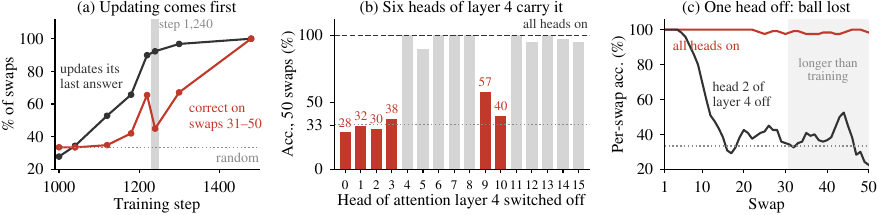}
\caption{\textbf{Tracking the aha moment.} (a)~Over training: share of swaps on which the frame-wise answer equals the model's own last answer updated by the current swap (black), and per-swap accuracy on swaps 31--50 (red). (b)~Accuracy on 50-swap games with one head of attention layer 4 switched off; any single head of attention layer 3 or 5 keeps at least 60\%. (c)~Per-swap accuracy with head 2 of attention layer 4 off.}
\label{fig:reuse}
\end{figure*}

\begin{figure}[t]
\centering
\includegraphics[width=\linewidth]{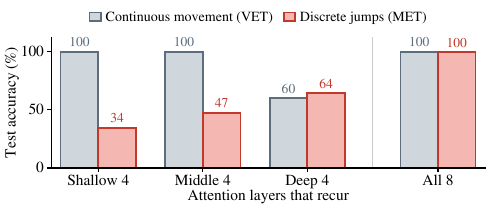}
\caption{\textbf{Shallow recurrence suits continuous motion; deep recurrence carries jumps further.} Accuracy after 2{,}500 steps with recurrence in only 4 of the 8 attention layers (shallow, middle, deep: 1--4, 3--6, 5--8).}
\label{fig:layers}
\end{figure}

\runin{An aha moment during training.}
Finally, we check how the training loss changes during training (\cref{fig:length}a--c).
We observe that the loss of \method on shell games stays on a plateau for about 1{,}200 steps and then drops to near zero within 300 steps, while its accuracy on unseen 50-swap games jumps from 35\% at step 1{,}000 to 86\% at step 1{,}500 and 99\% at step 2{,}500.
The loss of the baseline decreases slowly, and its accuracy stays at 30--38\% on 50-swap games for all 5{,}000 steps.
We call this jump the aha moment: length generalization emerges suddenly, not gradually over training.

\insight{Recurrence adds state transitions over time, not depth.}{Without recurrence, each swap costs an attention layer, so the ball is lost after about ten swaps. Recurrence gives every frame group its own state transition, and this ability emerges suddenly at an aha moment during training.}

\subsection{What happens at the aha moment?}
\label{sec:anatomy}
To understand this emergence, we focus on the aha moment (\cref{fig:reuse}).
First, we ablate the components of \method.
Second, we track the model before and after the aha moment.
Third, we probe inside attention layer 4.
Finally, we locate where the learned state is stored.

\runin{Ablating components.}
We first ask which components the aha moment needs (\cref{tab:ablation}a,\,b, \cref{fig:layers}).
Without frame-wise supervision, \method stays at chance, and without the recurrent KV cache it learns only the short cup games within 2{,}500 steps.
With recurrence in only 4 of the 8 attention layers, the shallow and middle layers learn the cup games, which move continuously, but not the shell games, which jump by a whole swap per frame group, and the deep layers learn neither fully (\cref{fig:layers}).
The aha moment thus needs frame-wise supervision and recurrence in all eight attention layers during training.

\begin{table}[t]
\centering
\small\setlength{\tabcolsep}{3.2pt}\renewcommand{\arraystretch}{1.0}
\caption{\textbf{Ablating components.} Accuracy (\%) after 2{,}500 steps, one change to \method per row: (a, b) in training, (c--e) only at inference; (a) is the baseline of \cref{tab:main} at step 2{,}500. Shaded: unseen lengths.}
\label{tab:ablation}
\begin{tabular}{@{}l cc cc@{}}
\toprule
Variant & VET & MET & \cellcolor{lightgray}Cup$\times$16 & \cellcolor{lightgray}Shell$\times$50 \\
\midrule
\oursrow
\textbf{\method} & 100 & 99.6 & 91.0 & 99.0 \\
\midrule
(a) no recurrent KV cache & 90.0 & 32.0 & 35.0 & 37.0 \\
(b) no frame-wise supervision & 36.0 & 33.8 & 35.0 & 32.0 \\
\midrule
(c) recurrent KV cache off & 34.0 & 34.8 & 40.0 & 30.0 \\
(d) keep last 8 frame groups & 98.0 & 99.6 & 97.0 & 91.0 \\
(e) keep last 4 frame groups & 38.0 & 91.8 & 37.0 & 55.0 \\
\bottomrule
\end{tabular}
\end{table}

\begin{figure*}[t]
\centering
\begin{minipage}[t]{0.465\textwidth}
\vspace{0pt}\centering
\includegraphics[width=\linewidth]{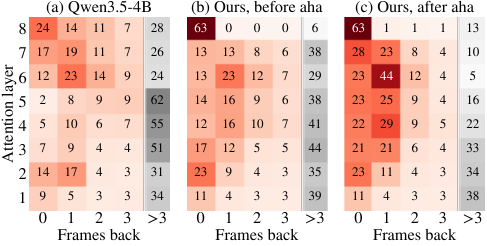}
\captionof{figure}{\textbf{Attention moves to the latest frame groups at the aha moment.} Share (\%) of image-token attention on the same frame group (0), the previous three, and older ones (gray), per attention layer, averaged over swaps 21--50 of 50-swap games.}
\label{fig:aha_attention}
\end{minipage}\hfill
\begin{minipage}[t]{0.51\textwidth}
\vspace{0pt}\centering
\small\setlength{\tabcolsep}{3.4pt}\renewcommand{\arraystretch}{1.0}
\captionof{table}{\textbf{Probing inside attention layer 4.} Accuracy (\%) of linear probes on held-out 50-swap games at three points of the layer: the swap is seen at its input, the last ball position is fetched by its attention, and the new position is then computed from both by its MLP.}
\label{tab:aha_probe}
\begin{tabular}{@{}ll ccc@{}}
\toprule
Model & Point in layer 4 & Swap & Last pos. & New pos. \\
\midrule
\multirow{3}{*}{\shortstack[l]{Qwen3.5-4B\\(fine-tuned)}} & Input & 100 & 38 & 34 \\
 & After attention & 100 & 34 & 36 \\
 & After MLP & 100 & 36 & 36 \\
\midrule
\multirow{3}{*}{\shortstack[l]{\method\\during aha}} & Input & 100 & 37 & 38 \\
 & After attention & 100 & \textbf{83} & 63 \\
 & After MLP & 100 & 85 & \textbf{80} \\
\midrule
\multirow{3}{*}{\shortstack[l]{\method\\after aha}} & Input & 100 & 37 & 38 \\
 & After attention & 100 & \textbf{100} & 95 \\
 & After MLP & 100 & 100 & \textbf{99} \\
\bottomrule
\end{tabular}
\end{minipage}
\end{figure*}

\runin{Tracking the aha moment.}
We then follow the frame-wise answers through training (\cref{fig:reuse}a).
The share of swaps on which the frame-wise answer equals the model's own last answer updated by the current swap rises from 28\% at step 1{,}000 to 92\% at step 1{,}240, before accuracy follows; the baseline stays at 41\%.

Six of the 16 heads of attention layer 4 carry this reuse: switching off any one of them drops the accuracy on 50-swap games to 28--57\%, and the ball is lost after the first swaps (\cref{fig:reuse}b,\,c).
In short, the model first learns to reuse its own last answer and update it with the current swap, through these six heads of attention layer 4.

\runin{Probing inside attention layer 4.}
If the update rule is in place by step 1{,}240, long-game accuracy should already be high, so we look inside attention layer 4 (\cref{tab:aha_probe}, \cref{fig:aha_attention}).
At the aha moment, attention layers 3--7 shift toward the current and previous frame groups; in attention layer 4, the share on the previous frame group rises from 16\% before to 29\% after the aha moment.

Probes read the current swap at the input of attention layer 4, the last position after its attention, and the new position after its MLP.
During the aha moment, its attention already fetches 83\% of the last positions, but its MLP still computes the new position for only 80\% of the swaps, against 99\% of the swaps after the aha moment.

Meanwhile, the number of swaps the model tracks before its per-swap accuracy falls below 80\% grows from about 13 at step 1{,}200 to 25 at step 1{,}240, 37 at step 1{,}400, 66 at step 1{,}480, and over 100 from step 2{,}000.
Copying the final MLPs of attention layer 4 and of the three GDN layers below it into the model of step 1{,}240 raises its per-swap accuracy on swaps 31--50 from 50\% to 82\%, while its final attention gives only 60\%.
The rule is thus in place before the jump, and the jump makes its update nearly exact.

\obsinf{The aha moment extends tracking.}{In the same 300 steps, the model starts to reuse its last answer and tracks from about 13 to over 100 swaps.}{The update rule forms first; accuracy jumps once the MLP of attention layer 4 computes the new position reliably.}

\begin{table}[t]
\centering
\small\setlength{\tabcolsep}{3.2pt}\renewcommand{\arraystretch}{1.0}
\providecommand{\dup}[1]{{\footnotesize\textcolor{black!85}{(+#1)}}}
\providecommand{\ddn}[1]{{\footnotesize\textcolor{black!40}{($-$#1)}}}
\providecommand{\rat}[1]{{\footnotesize\textcolor{black!85}{(#1$\times$)}}}
\caption{\textbf{General video understanding.} Accuracy (\%) on the Perception Test~\cite{patraucean2023perceptiontest} validation set by skill after fine-tuning; gray: question counts; parentheses: change vs.\ fine-tuned Qwen3.5-4B.}
\label{tab:pt}
\begin{tabular}{@{}l c cc@{}}
\toprule
\multirow{2}{*}[-3pt]{Question type} & \multicolumn{2}{c}{Qwen3.5-4B} & \method \\
\cmidrule(lr){2-3}\cmidrule(lr){4-4}
 & zero-shot & fine-tuned & fine-tuned \\
\midrule
Perception Test {\footnotesize\textcolor{black!55}{(2000)}} & 64.8 & 78.5 & 76.7\,\ddn{1.8} \\
\midrule
\multicolumn{4}{@{}l}{\textit{Advantage: the state after an action}} \\
\hspace{0.6em}Adversarial action {\footnotesize\textcolor{black!55}{(104)}} & 42.3 & 56.7 & 62.5\,\dup{5.8} \\
\hspace{0.6em}Object attributes {\footnotesize\textcolor{black!55}{(98)}} & 57.1 & 74.5 & 78.6\,\dup{4.1} \\
\hspace{0.6em}State recognition {\footnotesize\textcolor{black!55}{(166)}} & 62.7 & 76.5 & 79.5\,\dup{3.0} \\
\hspace{0.6em}Task completion {\footnotesize\textcolor{black!55}{(166)}} & 50.0 & 73.5 & 76.5\,\dup{3.0} \\
\midrule
\multicolumn{4}{@{}l}{\textit{Disadvantage: recalling an earlier moment}} \\
\hspace{0.6em}Sequencing {\footnotesize\textcolor{black!55}{(299)}} & 56.9 & 77.6 & 69.9\,\ddn{7.7} \\
\hspace{0.6em}Object recognition {\footnotesize\textcolor{black!55}{(252)}} & 70.2 & 85.3 & 78.2\,\ddn{7.1} \\
\hspace{0.6em}Action counting {\footnotesize\textcolor{black!55}{(138)}} & 38.4 & 71.7 & 66.7\,\ddn{5.0} \\
\hspace{0.6em}Feature matching {\footnotesize\textcolor{black!55}{(153)}} & 72.5 & 81.7 & 77.1\,\ddn{4.6} \\
\bottomrule
\end{tabular}
\end{table}

\begin{table}[t]
\centering
\small\setlength{\tabcolsep}{3.2pt}\renewcommand{\arraystretch}{1.0}
\providecommand{\dup}[1]{{\footnotesize\textcolor{black!85}{(+#1)}}}
\providecommand{\ddn}[1]{{\footnotesize\textcolor{black!40}{($-$#1)}}}
\providecommand{\rat}[1]{{\footnotesize\textcolor{black!85}{(#1$\times$)}}}
\caption{\textbf{Inference cost.} Seconds per answer for 8-fps cup videos on one RTX A5000 (ratio to Qwen3.5-4B in parentheses) and peak GPU memory, which is the same for both models.}
\label{tab:latency}
\setlength{\tabcolsep}{3pt}
\begin{tabular*}{\linewidth}{@{\extracolsep{\fill}}rr cc c@{}}
\toprule
\multicolumn{2}{c}{Video} & \multicolumn{2}{c}{Latency (s)} & \multirow{2}{*}[-3pt]{\makecell{Mem.\\(GB)}} \\
\cmidrule(lr){1-2} \cmidrule(lr){3-4}
Frames & Sec. & Qwen3.5-4B & \method & \\
\midrule
98 & 12 & 1.0 & 1.2\,\rat{1.2} & 9.1 \\
258 & 32 & 2.4 & 3.2\,\rat{1.4} & 10.0 \\
514 & 64 & 4.9 & 7.3\,\rat{1.5} & 11.4 \\
1{,}026 & 128 & 10.6 & 18.4\,\rat{1.7} & 14.4 \\
\bottomrule
\end{tabular*}
\end{table}

\runin{Locating the learned state.}
Finally, to locate the learned state, we use memory transfer: copying game B's recurrent KV cache into game A midway switches 98\% of A's later answers to B, those of the last frame group alone 83\%, and the state of all GDN layers none (\cref{fig:mechanism}c).
At test time, recurrence in attention layers 1--4 alone keeps a per-swap accuracy of 98.8\% on swaps 31--50, whereas switching it off only in attention layer 4 leaves 37.7\%.
Without recurrence, every game falls back to chance, and keeping only the last 8 frame groups lowers Shell$\times$50 from 99\% to 91\% (\cref{tab:ablation}c--e).
Together with the probes, which read the ball right after attention layer 4 (\cref{fig:mechanism}a), the learned state thus lives in the recurrent KV cache of attention layer 4, mostly of the latest frame group.

\insight{Every layer learns, one layer stores.}{All eight attention layers must recur to learn the rule; once it is learned, the state lives in attention layer 4 alone.}

\subsection{What does recurrence cost?}
\label{sec:cost}

\runin{General video understanding.}
We fine-tune Qwen3.5-4B with and without recurrence on the Perception Test~\cite{patraucean2023perceptiontest} training split (\cref{tab:pt}).
\method is 1.8 points lower overall: better at the state after an action, such as adversarial actions ($+$5.8), but worse at recalling an earlier moment, such as the order of events ($-$7.7): recurrence keeps the latest state but overwrites earlier details.

\runin{Inference cost.}
We time both models from loading a cup video at 8\,fps to the one-token answer (\cref{tab:latency}).
\method takes 1.2--1.7$\times$ the time of Qwen3.5-4B for videos of 12--128\,s, with the same peak memory.
The cost stays small because recurrence runs inside each attention layer: a frame group waits only for the previous group of the same layer, not for a pass through the whole model, as memory tokens do at 6.8$\times$ the training cost (\cref{tab:mechanisms}).

\insight{Recurrence trades detail for state.}{It keeps the latest state at the cost of details of earlier moments.}

\section{Conclusion}
\label{sec:concl}
We present \method, which adds a recurrent KV cache to a pretrained VLM so that its attention layers perform state transitions over time, without writing the state as text or calling external tools.
With a single answer token, \method achieves state-of-the-art accuracy on VET-Bench and MET-Bench and generalizes to games longer than in training, where fine-tuning without recurrence fails.
These results support our hypothesis that VLMs lack state transitions over time rather than data.
We further find that length generalization emerges suddenly at an aha moment, that each swap costs an attention layer without recurrence, and that the ball's state forms in the first four attention layers.
We hope that recurrent VLMs inspire more work on tracking the states of unobserved objects throughout a video.

\FloatBarrier

{\small
\bibliographystyle{abbrvnat}
\bibliography{paper}}

\end{document}